\documentclass[10pt,twocolumn,letterpaper]{article}

\usepackage{cvpr}
\usepackage{times}
\usepackage{epsfig}
\usepackage{graphicx}
\usepackage{amsmath}
\usepackage{amssymb}
\usepackage{authblk}

\usepackage{booktabs}       

\usepackage{color}
\usepackage{pifont}

\usepackage[pagebackref=true,breaklinks=true,letterpaper=true,colorlinks,bookmarks=false]{hyperref}

 \cvprfinalcopy 

\def\cvprPaperID{5129} 

\title{Spatial-Scale Aligned Network for Fine-Grained Recognition}
\ifcvprfinal
\fi

\author[1]{Lizhao Gao}
\author[2]{Haihua Xu}
\author[1]{Chong Sun}
\author[3]{Junling Liu}
\author[1]{Yu-Wing Tai}
\affil[1]{Tencent, \{leonardgao, waynecsun, yuwingtai\}@tencent.com}
\affil[2]{Megvii, xuhaihua@megvii.com}
\affil[3]{Peking University, liujunling@pku.edu.cn}
 
\begin{document}

\maketitle

\begin{abstract}
Existing approaches for fine-grained visual recognition focus on learning marginal region-based representations while neglecting the spatial and scale misalignments, leading to inferior performance.
In this paper, we propose the spatial-scale aligned network (SSANET) and implicitly address misalignments during the recognition process.
Especially, SSANET consists of 1) a self-supervised proposal mining formula with Morphological  Alignment  Constraints;
2) a discriminative scale mining (DSM) module, which exploits the feature pyramid via a circulant matrix, and provides the Fourier solver for fast scale alignments; 3) an oriented pooling (OP) module, that performs the pooling operation in several pre-defined orientations. Each orientation defines one kind of spatial alignment, and the network automatically determines which is the optimal alignments through learning.
With the proposed two modules, our algorithm can automatically determine the accurate local proposal regions and generate more robust target representations being invariant to various appearance
variances.
Extensive experiments verify that SSANET is competent at learning better spatial-scale invariant target representations, yielding the superior performance on the fine-grained recognition task on several benchmarks.

\end{abstract}

\section{Introduction}
Recognizing the fine-grained categories such as wild bird species \cite{WahCUB_200_2011}, automobile models \cite{KrauseStarkDengFei-Fei_3DRR2013}, is of great need in daily life.
The relevant technologies have alreay been applied in many internet products severing millions of users.
However, it is still a challenging task up to date for the inherent difficulties that the intra-class variance is sometimes larger than the inter-class one, which can be attributed to large pose, viewpoint and background changes.
It needs to extract the subtle visual details within subordinate classes under drastic appearance changes.
Hence the majority of efforts in the fine-grained community center on how to take full advantages of discriminative part localizations to learn marginal representations.
Some previous works  (both the traditional methods \cite{Gavves_2013_ICCV,xie2013hierarchical} and  CNN based methods \cite{lin2015deep,zhang2014part,zhang2014panda}) usually contain two steps:
1) object part localization by extra annotations; 2) feature extraction via the localized part region.
A main limitation for such methods is that they need large amounts of annotations for the part regions, which are not easy to be obtained though the existing computer vision algorithms.
As a result, detection-based methods and attention methods \cite{zheng2017learning,sun2018multi,fu2017look,zheng2019looking,yang2018learning,lin2015bilinear}  have been the focus of researches, as they no longer need extra part annotations.
These works try to learn a diverse collection of discriminative parts in an unsupervised manner through end-to-end model learning.
The learned part regions supplies complementary but vital information lost by the backbones throughout a chapter of pooling operations.
Recently, there are two different pathways of part-based methods in general. One pathway is to use soft attention mechanisms, \emph{e.g.}, MACNN \cite{zheng2017learning} utilizes a channel attention module to distinguish different parts and then jointly consider the recognition results from different parts for more robust estimation.
Another way is to use "hard" attention, namely selecting appropriate anchors or sub-windows to highlight the local regions with semantic importance.

Despite achieving promising progress, there are several critical issues left unsolved: 1) owing to the lack of annotations, both the attention-based and anchor-based methods fail to regress the accurate semantically informative regions (e.g., head of a bird);
2)  the targets to be recognized have diverse poses, rotations, viewpoints causing large intra-class variance.
Previous methods like \cite{jaderberg2015spatial,zhang2014part,lin2015deep,gavves2013fine} focus on addressing certain misalignment using complicated
design while failing to cover all the conditions.

To address the above challenges, we propose the SSANET, which focuses on learning accurate region proposals as well as robust target representations being robust to scale, orientation, rotation changes. Our contributions are summarized as follows:
\begin{itemize}
\item We propose the SSANET to learn better accurate and informative part features via self-supervised training, including Morphological Alignment Constraints for coarse part localization, a Discriminative Scale Mining module for scale alignment and an oriented pooling module for translation, orientation and rotation alignment.
\item 
We conduct extensive experiments on three challenging datasets (CUB Birds, Stanford Cars and FVGC aircraft), and demonstrate that our SSANET outperforms published region-based methods. Our network can be embedded into the existing convolutional architectures, making the networks be robust to spatial, scale and rotation changes.
\end{itemize}

The rest of the paper is organized as follows: Sect. 2 describes and discusses the works that are related to our approach, followed by the methodology and implementation of the SSANET in Sect. 3. Finally, we evaluate the proposed method on four widely-used datasets in Sect. 4.

\section{Related Work}

\subsection{Fine-Grained Classification}
The task of fine-grained visual recognition is to distinguish between animal species or several rigid bodies, like cars and aircraft. Since the inter-class differences are subtle and intra-class differences are variant due to distracted background context, there is been research in improving the attentional and localization techniques based on CNN.
Early works \cite{xie2013hierarchical,zhang2014part}
leverage manual part annotations to achieve promising results. Lately, more practical methods are proposed focusing on localizing object parts without requiring expert but expensive annotations. Attention-based CNNs like \cite{zheng2017learning,sun2018multi} use channel-wise attention mechanism to extract part information.
There is a current trend, which fuses the features of complementary parts with the feature of the input image. Part-based Region CNNs such as \cite{yang2018learning} navigates the most informative part regions to o strength the classifier.
\cite{fu2017look} learns discriminative region attention and region-based feature representation via combining attention proposal network with region-based classifier sequentially.
Furthermore, a series of parallel CNNs \cite{gao2016compact,lin2015bilinear,cui2017kernel} extract bilinear texture feature to stabilize spatial invariance.
\cite{jaderberg2015spatial} takes advantage of affine transformation for better alignment.
\cite{NIPS2018_7344} obtains the minimum information by the principle of Maximum-Entropy.
Here our proposed method makes the most use of part information organically. Focusing on aligning features among part regions in translation, scale, and rotation, our method can obtain competitive performance only utilizing predictions of part regions.

\subsection{Feature Alignments}

Many tasks such as fine-grained recognition, incorporate a variety of dispersive, subtle but vital details so need attention and detection mechanisms for discriminative features extraction and selection.
Recently, some works \cite{zhang2014part,lin2015deep,yang2018learning,fu2017look,zheng2017learning} manipulate the data by separating images into several parts to learn translation invariance. However, owing to lacks of annotations, these works with selective attention suffer from inaccurate detection boxes and pose misalignment.
One significant step is STN \cite{jaderberg2015spatial} that have shown spatial manipulation by data-driven enable to learn some degree of invariance to warps.
Here, our method not only focuses on achieving more accurate localization but also attempt to seeking unities of features in common, instead of eliminating the otherness among samples by an affine transformation.

\section{Our Approach}

\begin{figure*}[htbp]
\includegraphics[width=\textwidth]{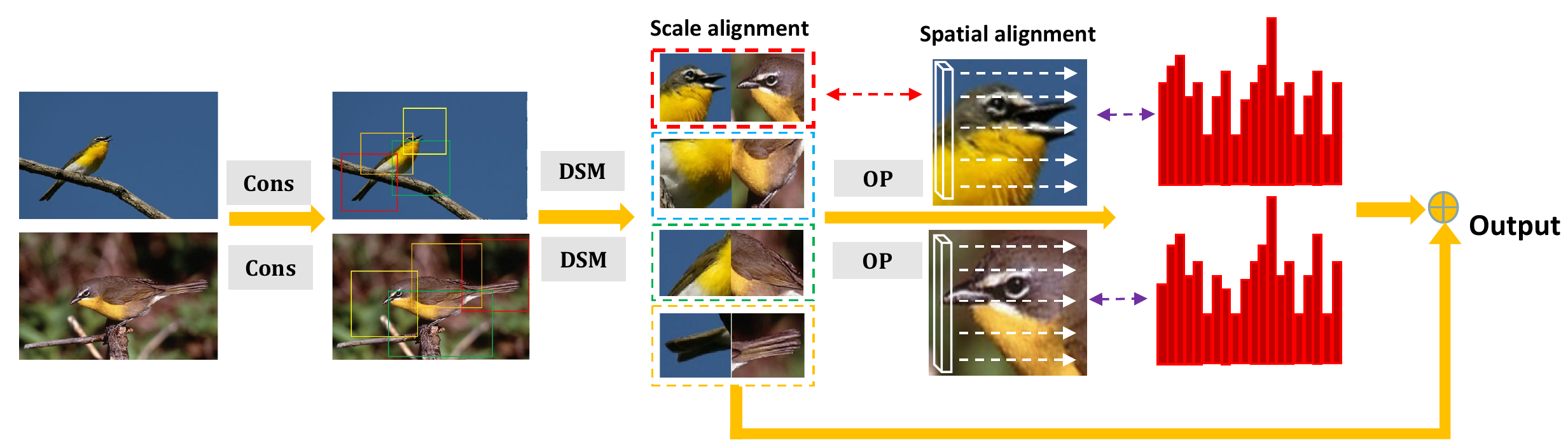}
\centering
\caption{The framework of SSANET. Starting from bottom-up region proposals, we harness
morphological alignment constraints and NMS to select top-N proposals. Then utilizing the DSM module to obtain robust scale estimation and OP module to obtain patial-invariance representations. The recognition results of both modules are fused for the ultimate prediction. Best viewed in color.}
\label{fig main}
\end{figure*}


In this section, we introduce our SSANET, which enables to align object features before fine-grained recognition.
An overview of the proposed SSANET is presented in Figure \ref{fig main}.
Our SSANET contains three key modules: 1). Morphological Alignment Constraints (MAC) for joint proposal generation and semantic level alignment, 2). Discriminative Scale Mining (DSM) module for accurate localization and 3). Oriented Pooling (OP) modules for spatial information alignment in partial branches.
Given an input image, we first feed it into several convolutional layers to extract feature tensors for the following processes.
Then we generate several part proposals via Morphological Alignment Constraints and define a loss function to select the top-K most informative region proposals.
All the proposals are passed into the Discriminative Scale Mining module to simultaneously perform multi-scale estimation and scale alignment.
The proposals with the aligned scale are further input the OP module to provide the spatial-invariant target recognition.
The recognition results from both the DSM module and the OP module are summed together as the final recognition results.


\subsection{Morphological Alignment Constraints}


SSANET generates regions by anchor-based methods, like \cite{yang2018learning}. We leverage $k$ scales and $k$ aspect ratios to yield $whk^2$ candidate informative proposal regions at each sliding position in convolutional feature map of size $w \times h$. We harness non-maximum suppression (NMS) to reduce region redundancy, and take the top-$K$ regions $A$ = \{$A_1,A_2, \dots, A_K$\} enabling discriminative part localizations.
We exploit an additional sub-network for further informative proposal selection.
The selected proposals are expected to obey the following two criteria: (1) the selected proposal region should have enough
information entropy;  (2) the features extracted in the regions with semantically similar representations (\emph{e.g.}, "heads", "legs") should be similar.
The first criterion ensures that the discriminative regions are selected for better recognition results, while the second criterion ensures that different categories select the proposal
regions with semantic correspondence. In the recognition process, we feed the proposals into the sub-network and sort them based on the network output, the top-$K$ proposals are selected for the following procedures.

\begin{figure}[htbp]
\includegraphics[width=0.36\textwidth]{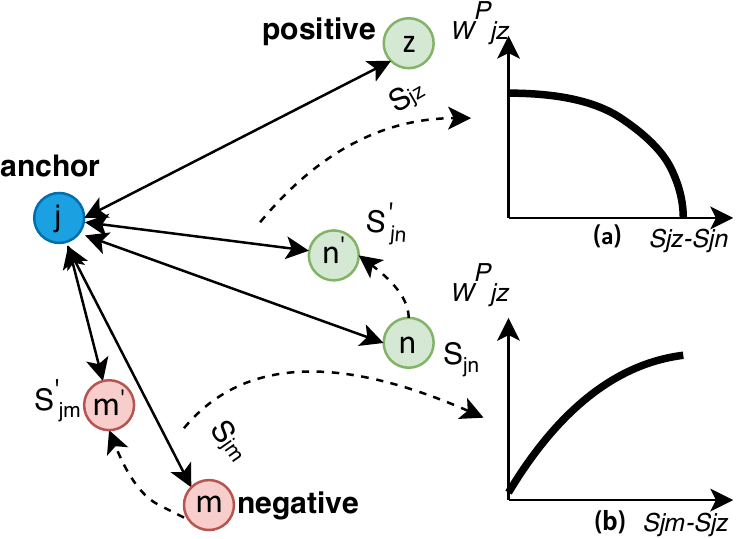}
\centering
\caption{The proposed $\mathcal{L}_{\rm m}$ is able to adjust the weights among different training pairs.
Let $S_{jz}$ and $S_{jn}$ denote the cosine scores of the positive training pairs, and $S_{jm}$ denote the negative similarity score. The weight $W_{jz}^P$ will be assigned to small values in two situations: (1) $S_{jz}$ is larger than similariy scores of other positive training pairs (\eg, $S_{jn}$); (2) $S_{jz}$ is obviously larger than similarity scores of negative training pairs (\eg, $S_{jm}$).
By this means, our method is able to implicitly ignore the easy positive samples in the training process.}
\label{fig mac}
\end{figure}

Given $B$ samples in one mini-batch, we divide each sample into $K$ regions and introduce a loss function for the proposal selection sub-network based on criterion 2.
We use $f_{i,j},i\in\{1,...,B\}, j \in \{1,...,K\},$ to denote the vectored feature extracted for the $j$-th region of the $i$-th sample. In our work, we constrain that the
regions with the same index to have semantic correspondence. For each training batch, we define our loss for proposal selection as follows:
\begin{equation}
\mathcal{L}_{\rm total} = \sum_{i=1}^B ( \sum_{j=1}^K \mathcal{L}_{\rm CE}({\cal FC}(f_{i,j}))+ \lambda \sum\limits_{\xi = 1,\hfill\atop\tau = 1}^{B}\mathcal{L}_{\rm m}(f_\xi, f_\tau)
\label{equ:1}
\end{equation}

\begin{equation}
{{\cal L}_{\rm m}}(f_\xi, f_\tau) = \frac{1}{K}\sum_{j=1}^K \log (1 + \sum\limits_{n \in K, \hfill\atop n=j} \sum\limits_{m \in K, \hfill\atop m \ne j} e ^{-S_{jn}+S_{jm}})
\label{equ:mac}
\end{equation}
where ${\cal FC}(f_{i,j})$ denotes one fully-connected layer whose output is the prediction probability of different species.
$\mathcal{L}_{\rm CE}$ is cross-entropy loss whose label is identical to the sample label and it is based on criteria 1.
$\mathcal{L}_{\rm m}(f_\xi, f_\tau)$ is defined based on criteria 2, $\lambda$ is a hyper-parameter.
We assume $S_{jn} = f_{\xi,j}^\top{f_{\tau, n}}$ as the cosine similarity between the $j$-th region and the $n$-th region of samples $\xi$ and $\tau$.

The derivative for $\mathcal{L}_{\rm m}$ with respect to model parameters $\theta$ at the $t$-th iteration can be calculated as follows:
\begin{equation}
\dfrac{\partial {\cal L}_{\rm m }}{\partial \theta} \mid_t = \sum_{j=1}^K \sum_{z=1}^K \dfrac{\partial {\cal L}_{\rm m}}{\partial S_{j,z}} \dfrac{\partial S_{j,z}}{\partial\theta} 
= \sum_{j=1}^K \sum_{z=1}^K  \omega_{jz} \dfrac{\partial S_{i,j}}{\partial\theta} 
\end{equation}
where $\omega_{jz}$ can be regarded as a constant weight in the gradient w.r.t. $\theta$. 
When the region index $j=z$, the corresponding $w_{jz}$ can be computed as:
\begin{equation}
\omega_{jz}^{P}=  \dfrac{\partial {\cal L}_{\rm m}}{\partial \left ( S_{jz} \right )} \mid_t  = \frac{1}{\frac{1}{\sum\limits_{m \ne j}e^{S_{jm}-S_{jz}}}+\sum\limits_{n=j}e^{S_{jz}-S_{jn}}}
\label{equ:ppair}
\end{equation}

When the region index $j\ne z$, the corresponding $w_{jz}$ can be computed as:
\begin{equation}
\omega_{jz}^{N}= \dfrac{\partial {\cal L}_{\rm m}}{\partial \left ( S_{jz} \right )} \mid_t  = 
\frac{1}{\frac{1}{\sum\limits_{n=j}e^{S_{jz}-S_{jn}}}+\sum\limits_{m \ne j}e^{S_{jm}-S_{jz}}}
\label{equ:npair}
\end{equation}

According to Eq.\ref{equ:ppair}, \ref{equ:npair}, by exploiting the MAC module, our method can dynamically determine the weights for different training pairs. Two kinds
of training pairs will be assigned larger absolute value of weights to, \ie, the positive training pair with small cosine similarity and the negative training pair with large cosine similarity. Through this mechanism, our method is able to automatically determine the valuable training samples (hard positive samples and hard negative samples). In Figure~\ref{fig mac}, we present an toy example on how this is achieved.

\begin{figure}[htbp]
\centering
\includegraphics[width=0.45\textwidth]{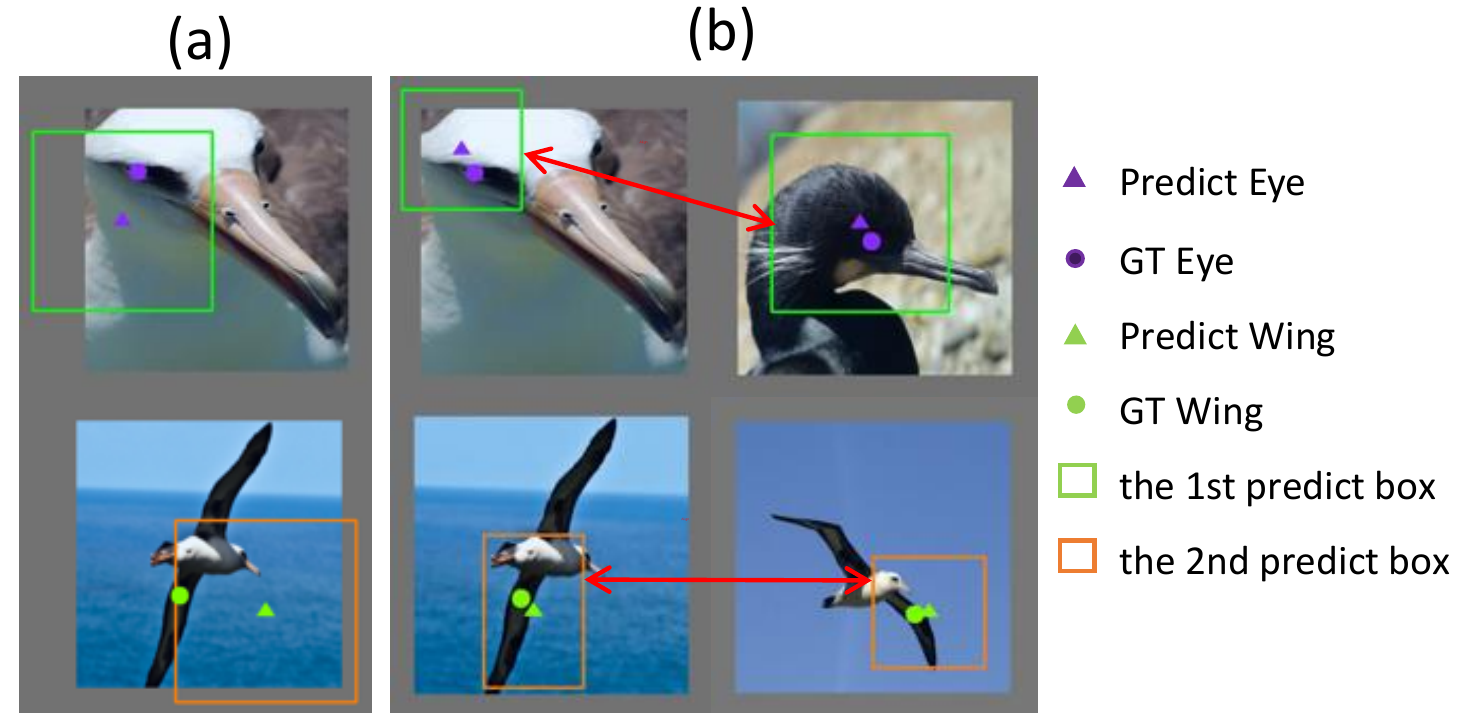}
\caption{Qualitative results of the DSM module for the CUB dataset. Dots mean GT center annotated and triangle points mean the center of our proposals. Green boxes in the $1st$ row focus on the eye of birds while orange boxes in the $2nd$ row focus on the wing of birds. We can observe that proposals generated by our DSM module (b) are more similar to GT than NTSNet (a), which is one critical step to obtain superior performance.}
\label{multiscale_result1}
\end{figure}

This module only performs an initial and coarse alignment which is not accurate (see the left section in Figure \ref{multiscale_result1}) with severe spatial and scale misalignments. The spatial-scale alignments are further considered in the following sections.

\begin{figure}[htbp]
\centering
\includegraphics[width=0.46\textwidth]{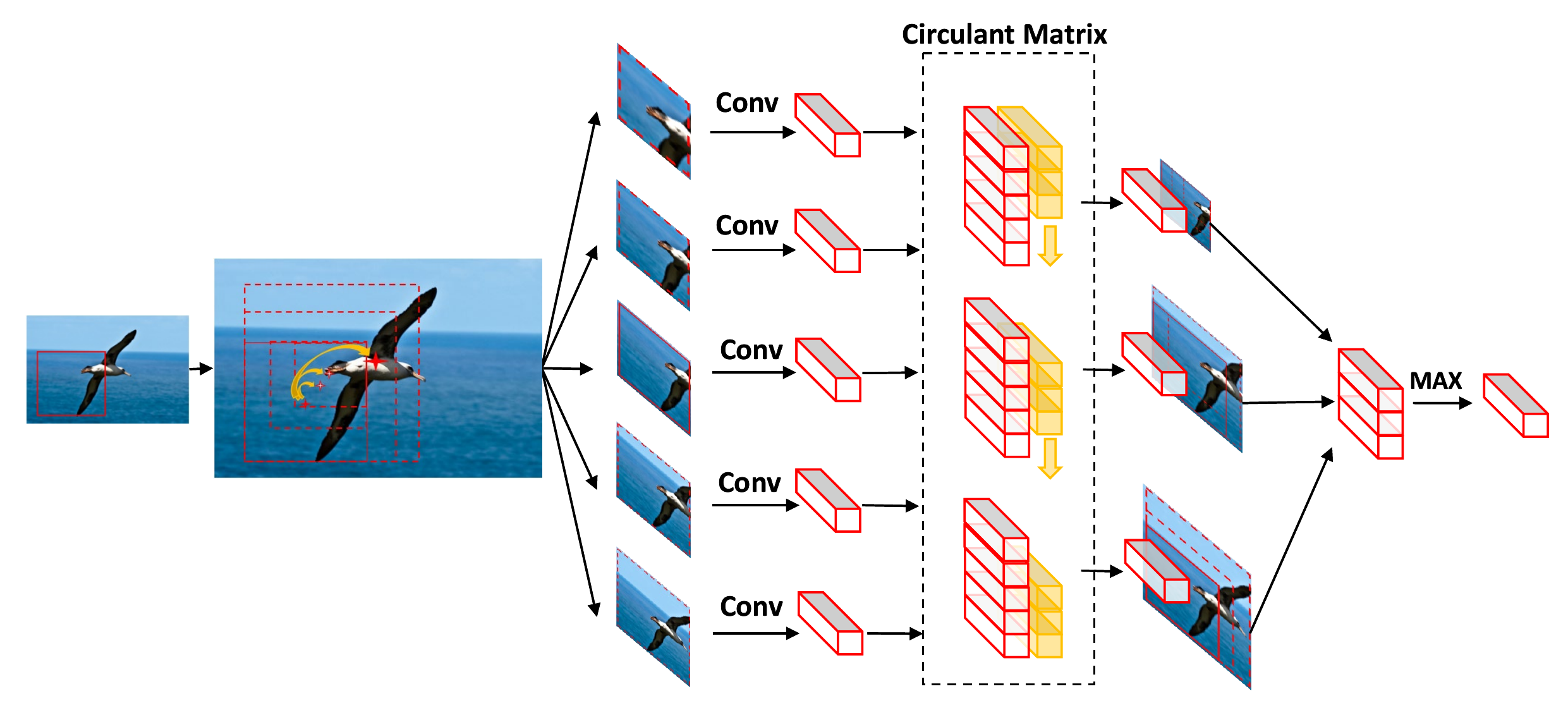}
\caption{The framework of the DSM module.  For one feature map $f^{h \times w}$,  we obtain $S$ candidates via expanding and shrinking, and then flatten $f^{h \times w}$  to compose the circulant matrix $C_k$ where $k=h \times w \times C$. Then we compute the response $y_i$ according to Eq. \ref{equ:scale_alignment}.}
\label{fig dsm}
\end{figure}
\subsection{Discriminative Scale Minining Module}
Since no annotated local regions are available, the previous generated local proposals may be not accurate causing scale misalignments.

In this section, we exploit a circulant matrix to simultaneously perform scale alignment and multi-scale estimation. Let the tube $A_i = [\phi_i, \psi_i,\eta_i, \gamma_i]$ denote the state of an object proposal, where $\phi_i$ denotes the extracted feature tensor for the $i$-th proposal, $\psi_i$ and $\eta_i$ are the two vertexes of the rectangular proposal region $i$,
  $\gamma_i = (w_i, h_i)$ denotes the height and width of the proposal.
 In our algorithm, we first compute the center coordinate for all the selected proposal regions, and determine $\psi_i$ as the vertex farthest to the center and $\eta_i$ as the vertex nearest to
 the center. We expand or shrink the original proposal region, and generate $S$ (set as an odd) proposal candidates.
 Fixing the vertex $\psi_i$, we expand the scale of the proposal obtaining $\frac{S + 1}{2} $ candidates as $A_i = [\phi _i^n,{\psi _i},{\gamma _i}{\alpha ^{ n - {\frac{S - 1}{2}}}}]_{n=1}^{ {\frac{S + 1}{2}}}$, where we use $\alpha$ to denote the scale factor for candidate generations.
Similarly, we also shrink the proposal region fixing $\eta_i$, and obtain the candidates as $B_i = [\phi _i^n,{\eta _i},{\gamma _i}{\alpha ^{n -  {\frac{S - 1}{2}} }}]_{\frac{S + 1}{2}}^S$.
 For simplicity, we only consider the response activation for one category as an example to present the algorithm, the conclusion can be extended to the multi-category recognition case.
 Given multi-scale candidates, we can compute the response considering multiple scales as ${y_i} = \sum\limits_{n = 1}^S { {w_i^n}^\top\phi _i^n } $, which is widely used in many multi-scale based neural networks. However, directly computing the response considering different proposal scales is suboptimal as scale sizes of different candidates are not well aligned.
Since the scale of each proposal may not be accurate, the filter ${w_i^n}$ and $\phi _i^n$ are not well aligned for different targets. In this work, we exploit the circulant matrix to implicitly perform scale alignment while computing the response $y_i$.
The circulant matrix is essentially a Toeplitz matrix, which has been widely used in signal processing and several computer vision tasks (\emph{e.g.},~\cite{danelljan2016discriminative}).  We use ${C}_k$ to denote the circulant matrix constructed via $\phi _i^1(k), \phi _i^2(k), ..., \phi _i^S(k)$, where $\phi _i^n(k)$ denotes the $k$-th element of feature vector $\phi_i^n$:
\begin{equation}
{C}_k = \left[ {\begin{array}{*{20}{c}}
{\phi _i^0(k)}&{\phi _i^1(k)}& \cdots &{\phi _i^S(k)}\\
{\phi _i^S(k)}&{\phi _i^0(k)}& \cdots &{\phi _i^{S - 1}(k)}\\
 \vdots & \vdots & \ddots & \vdots \\
{\phi _i^1(k)}&{\phi _i^2(k)}& \cdots &{\phi _i^0(k)}
\end{array}} \right]
 \end{equation}
 
 Exploiting ${{{C}}_k}|_{k = 1}^K$, we compute the response $y_i$ considering scale alignments as
 \begin{equation}
 {y_i} = \max \sum\limits_{k = 1}^K {{C_k}{w_i^k}} ,
 \label{equ:scale_alignment}
 \end{equation}
where ${w_i^k} = [w_i^0(k), w_i^1(k), \cdots, w_i^S(k)]^\top$.
  In Eq.~\ref{equ:scale_alignment}, we consider different scale combinations exploiting various permutations of base vector $\psi _i^k = [\phi _i^0(k),\phi _i^1(k), \cdots ,\phi _i^S(k)]$. According to the property of the circulant matrix, Eq.~\ref{equ:scale_alignment} can also be rewriten as
 ${y_i} = \sum\limits_{k = 1}^K {{ \mathcal{ F}^{ - 1}}({{(\mathcal{ F}\psi _i^k)}^C} \odot (\mathcal{ F}w_i^k)}) $, where $\mathcal{ F}$ is the Fourier matrix, $(.)^C$ denotes the conjugate operation of a complex-valued vector. It has the computation complexity of $O(KS{\rm log}_2S)$, which is computation efficient.
Suppose $\delta_i $ is the error term corresponding to $y$, we need to compute the gradient of $w_i^k$ and the error term of $\phi_i^n$ for this module during backpropagation to enable the end-to-end training of the proposed network. Given $\delta_i$, the gradient with respects to $w_i^k$ can be computed as
\begin{equation}
\frac{{d \mathcal{L}. }}{{dw_i^k}} = {\delta _i}{\phi_i^*}^\top
\end{equation}
where $\phi_i^*$ is the permuted vector of $\psi _i^k$ which is selected through Eq.~\ref{equ:scale_alignment}. The error term of $\phi_i^*$ can be similarly computed as $\frac{{d\mathcal{L}}}{{d\phi _i^*}} = {\delta _i}{w_i^k}^\top$.

\subsection{Oriented Pooling Module}
Spatial misalignments lead to inferior performance in many computer vision tasks, such as person re-identification, face recognition, to name a few. It can be caused by many complex reasons, \emph{e.g}., camera viewpoint changes, drastic pose changes of the target object.
As a successful alternative, the global average pooling method fuses the per-channel network responses of different spatial locations into one and partially address the displacements. However, it comprises the discriminative power as the spatial information is greatly compressed.

In this work, we improve the global average pooling operation by proposing the new oriented pooling module.
Different from the global average pooling method, we perform the pooling operation on several predefined orientations to address different kinds of misalignments.
To capture the intra-category consistent features, we use four pooling orientations, \emph{i.e.}, the horizontal and vertical max pooling on the original and rotated feature maps.
Specially, given a feature map ${ X} \in \mathbb{R}^{H\times W \times C}$, we first divide it into $N_H\times N_W$ overlapping sub-patches with size $h\times w$. We reshape the feature tensor in each sub-patch as ${ x}_{i,j}\in \mathbb{R}^{K}$, $i\in\{1,...,N_H\}, j\in\{1,...,N_W\}$, $K = h\times w\times C$.
We use ${p_i}$ to denote the operation $\mathop {\max }\limits_{1 \le j \le {N_w}} \left\{ {{x_{i,1}},{x_{i,2}},...,{x_{i,{N_w}}}} \right\}$ and use ${q_j}$ to denote $\mathop {\max }\limits_{1 \le i \le {N_H}} \left\{ {{x_{1,j}},{x_{2,j}},...,{x_{{N_H},j}}} \right\}$.
The pooling operation based on the four pre-defined orientions can be described as:

\begin{equation}
{\Gamma _h} = {[{p_1}^ \top ,{p_2}^ \top ,...,p_{{N_H}}^ \top ]^ \top }
\end{equation}

\begin{equation}
{\Gamma _v} = {[{q_1}^ \top ,{q_2}^ \top ,...,q_{{N_W}}^ \top ]^ \top }
\end{equation}

\begin{equation}
{{\Gamma '}_h} = {[{p_{{N_H}}}^ \top ,{p_{{N_H} - 1}}^ \top ,...,p_1^ \top ]^ \top }
\end{equation}

\begin{equation}
{{\Gamma '}_v} = {[{q_{{N_W}}}^ \top ,{q_{{N_W} - 1}}^ \top ,...,q_1^ \top ]^ \top }
\end{equation}

In the recognition process, only one pooled feature vector is selected, thus we obtain the activation considering different pooling orientations as

\begin{equation}
y = \max ({{{w}}^\top}{\Gamma _h},{{{w}}^\top}{\Gamma _v},{{{w}}^\top}{{\Gamma '}_h},{{{w}}^\top}{{\Gamma '}_v})
\end{equation}
where $w$ is classifier weight which is indeed the weight matrix in the fully connected layer. By exploiting the proposed oriented pooling features, the target object obtained with different viewpoints and poses can be better aligned. We address the rotation issue via the proposed OP module. Suppose we have a $2\times 2$ gray-scale image $\left[ {\begin{array}{*{20}{c}}{\rm{a}}&b\\c&d
\end{array}} \right]$ and the rotated version $\left[ {\begin{array}{*{20}{c}}c&a\\
d&b\end{array}} \right]$, according to OP operation, we can obtain the consistent target representations for images with different rotation angles, i.e., ${\Gamma _v} = [\max (a,c),\max (b,d)]^\top$ for original image and ${\Gamma _h} = [\max (a,c),\max (b,d)]^\top$ for the rotated image. The OP module provides consistent target representations against different rotation angles. What is more, the spatial information is better retained than the global average pooling method.

\subsection{Training SSANET}
We train the network with 3-stage strategy. At the first stage, we train the DSM module initialized with ILSVRC2012 pre-trained model to achieve better localization. At the second stage, we train the OP module using the proposals generated by the DSM module in the first stage fixing the parameters of the CNN feature extractor and the DSM module. Finally, keeping the parameters of DSM and OP fixed, we fine-tune part-net for several iterations. As an alternative, we also implement the one-stage training strategy that jointly trains the entire network. We find that the OP module does not work
well in the joint training variant, which only has marginal improvement compared to the baseline network.

\section{Experiments}

\subsection{Experimental setup}
\paragraph{Datasets} We conduct experiments on three challenging fine-grained datasets, namely Caltech-UCSD Birds \cite{WahCUB_200_2011}, Stanford Cars \cite{KrauseStarkDengFei-Fei_3DRR2013} and FGVC Aircraft \cite{maji2013fine}. 

\begin{table}
\small
  \caption{Rotation robustness evaluation of different methods on the CUB-200-2011 dataset.}
  \label{rotation}
  \centering
  \begin{tabular}{lll}
  \hline
    \toprule
    Method        & Original input & Rotated input \\
    \midrule
  ResNet-50  &84.5   &57.3  \\
  NTSNet \cite{yang2018learning}   &87.5   &61.9  \\
  SSANET     &88.5   &\boldmath${65.9}$ \\
    \bottomrule
  \end{tabular}
 \end{table}
 
\begin{table}
\small
  \caption{Scale robustness evaluation of different methods on the CUB-200-2011 dataset.}
  \label{shrink}
  \centering
  \begin{tabular}{lll}
  \hline
    \toprule
    Method        & Original input &  Reduced input \\
    \midrule
  ResNet-50  &84.5   &76.0  \\
  NTSNet \cite{yang2018learning}    &87.5   &82.6  \\
  SSANET     &88.5   &\boldmath${83.2}$ \\
    \bottomrule
  \end{tabular}
 \end{table}

\begin{table}
\small
  \caption{Ablation analysis of our method on CUB-200-2011. Three sections are divided by the horizontal separators. 'P' in parentheses denote proposals.}
  \label{ablationstudy}
  \centering
  \begin{tabular}{ll}
  \hline
    \toprule
    Method        & Top-1 \\
    \midrule
    ResNet-50              & 84.5  \\
    ResNet-50(P)    & 86.8  \\
    ResNet-50(P) + MAC       & 87.3  \\
    \midrule
    ResNet-50(P) + MAC  + OPM           & 88.1  \\
    ResNet-50(P) + MAC  + DSM          & 88.3  \\
    \midrule
    ResNet-50(P) + MAC + OP + DSM  & \boldmath${88.5}$ \\
    \bottomrule
  \end{tabular}
\end{table}


\paragraph{Implementation Details}
We use Momentum SGD optimizer at both 3 stages and the learning rate is multiplied by 0.1 after 60 epochs. The weight decay is set to 1e-4. At the first stage, the initial learning rate is set to 0.001 for all training parameters. And at the second stage, the learning rate is set to 0.001 for parameters of the OP module and 3 times lower for parameters in the global fully-connected classifier.
On the Stanford Cars dataset, the scale size is set to 512 and the crop size is set to 448. The learning rate is multiplied by 0.1 after 30 epochs when training the OP module. On the FGVC Aircraft dataset, we resize the input images into 448 straightly. The weight decay is set to 5e-4 and the batch size is set to 24. The hyperparameter $\lambda$ is empirically set to 0.5 as it achieves good performance. In addition, we try several values (e.g. 0.1, 0.3, 1), which achieve 88.4\%, 88.5\%, 88.5\% respectively. Our SSANET is robust to $\lambda$ because the network weights of the backbone are frozen when we train the DSM module and OP module. The number of parts is fixed for all the datasets in this paper. We find that the experimental results only have marginal improvements when the part number is larger than 4, thus it is set as 4 in the implementation. 

Our method can achieve real-time inference with a Tesla P40 GPU (50 ms). It takes about 40ms during the DSM module while the computation time for the OP module can be ignored. It is a bit time-consuming for the DSM module owing to that the feature extraction network for multi-scale candidates is not shared. There are great potentials for the method to be accelerated when the feature computation of various candidates is shared (ROI pooling can be used for this purpose).

\begin{figure}[htbp]
\centering
\includegraphics[width=0.25\textwidth]{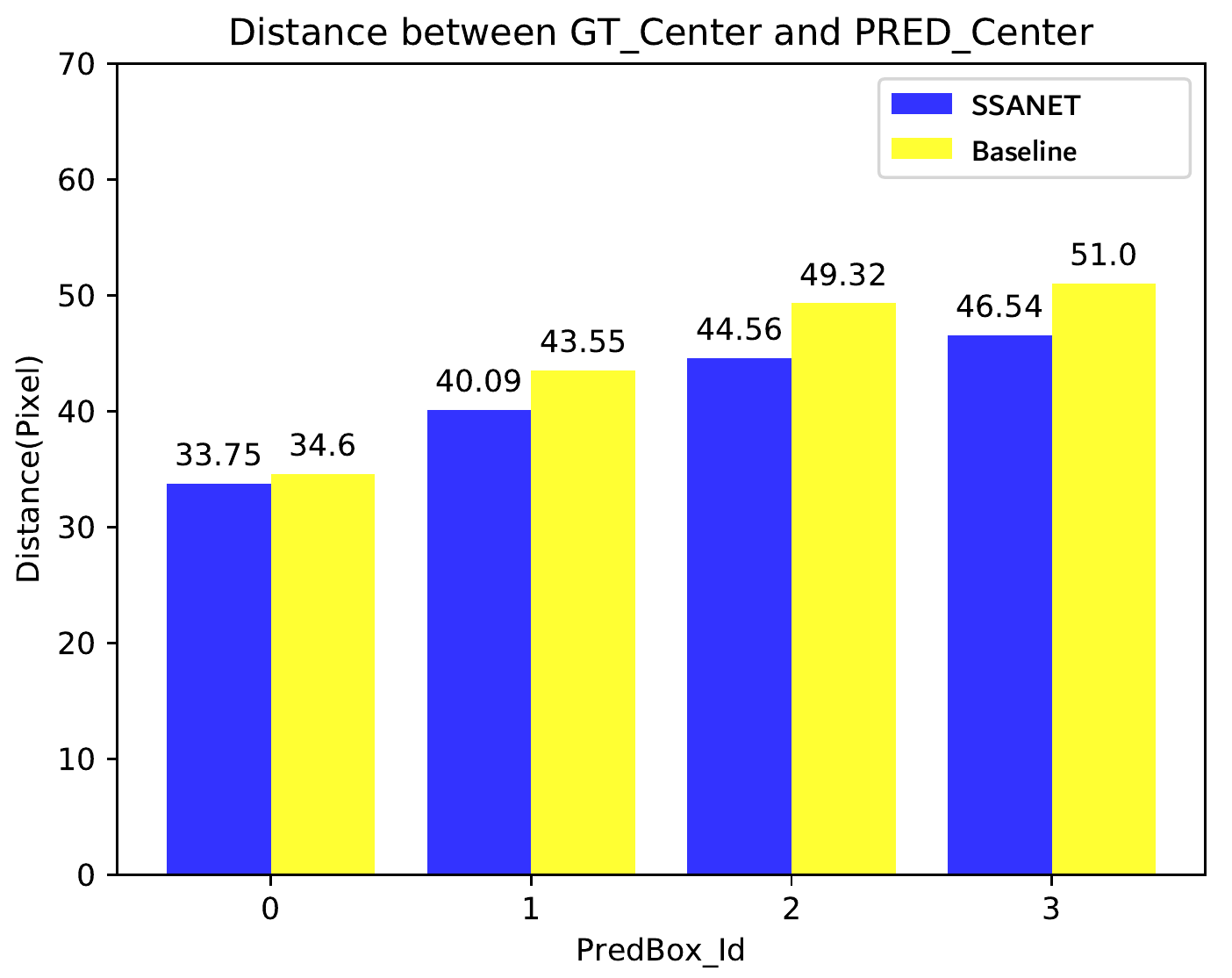}
\caption{Point Distance between the generated proposal center and ground-truth key points for both SSANET and the baseline method.}
\label{distance}
\end{figure}

\subsection{Ablation Studies}
In this section, we conduct ablation studies on the CUB birds dataset to validate the effectiveness of the OP module, the DSM module and the integration module.
We exploit ResNet-50 as our backbone as it has been validated to be effective in various computer vision tasks.
Performance of different variants of the proposed method is presented in Table \ref{ablationstudy} for comparison.
It is interesting to observe that by solely exploiting the ResNet-50 network, we obtain a 84.5\% top-1 accuracy, which is comparable to many state-of-the-art methods.

\paragraph{effectiveness of MAC}
Here we validate the effectiveness the proposals generated with the Morphological Alignment Constraints. We use ResNet-50 to denote the method only exploiting ResNet-50 backbone for target recognition.
ResNet-50(P) is the variant considering proposals, and ResNet-50(P) + MAC denotes the method further considerint the MAC module. 
Both the proposals and the MAC module leads to better recognition results and the MAC module leads to a $0.5\%$ performance improvement on the CUB 200-2011 dataset.

The generated proposals (without the two alignment modules) improve the backbone method by 2.8\%, and this shows that
part regions can provide good complementary for the holistic recognition task.
In this section, we take this method as baseline for the following experiments to further analyze the effect of sub-network settings.

\paragraph{effectiveness of OP module}
The OP module tries to obtain invariant target representations under various pose and scale changes. As is described in Sect. 3.3, we perform the pooling operation on the patch level.
In our implementation, we set $h = w  = 3 $ and $N_H = N_W = 3$ in the OP module. From Table \ref{ablationonopm}, it can be seen that the OP module improves the baseline by 0.8\%. 




\paragraph{effectiveness of DSM module}
In the DSM module, both multi-scale estimation and scale alignment is implicitly performed. To validate the effectiveness of the proposed DSM module, three experiments are conducted in different settings.
The first setting \emph{DSM(fixed)} fixes the combinations of 5 different scales (One original, two expanded, two shrank), the second setting \emph{DSM(1)} considers 5 scales of proposals and only exploits one scale for target recognition, and the third setting \emph{DSM} is the implementation of the proposed algorithm.

Table \ref{ablationondsm} shows the experimental results of different settings. Compared to our algorithm, \emph{DSM(fixed)} only considers the multi-scale estimation, whilst ignoring the scale alignment process, \emph{DSM(1)} considers the scale alignment and ignores the the multi-scale estimation. Our method \emph{DSM(1)} jointly considers the multi-scale estimation and scale alignment, and thus achieves superior performance. This experiment demonstrates the effectiveness of the proposed DSM module. To prove the effectiveness of the DSM module further, we conduct experiments adding SPPNet and FPN on the ResNet-50 + MAC with the same settings as \cite{he2015spatial} and \cite{lin2017feature} respectively.
As shown in Table \ref{ablationondsm}, the DSM module outperforms both SPPNet and FPN by 0.9\%, 0.7\% respectively.
Furthermore, 15 key points (GT) are provided for each image in the CUB-2011 dataset. We select the top 4 proposals for the SSANET and the baseline respectively and compute the distance between the proposal center and the nearest key point. The average distance corresponding to the 4 proposals of different methods are provided in Fig. \ref{distance}, which shows that the DSM module estimates more accurate proposal for target recognition.

The previous experiments have validated that both the OP and DSM modules improve the baseline method to some extent. By combing these two modules in the same network, further performance
gain can be obtained. This is validated by comparing ResNet-50 + MAC + OP + DSM (\emph{i.e.}, SSANET) with ResNet-50 + MAC + OP and ResNet-50 + MAC + DSM. SSANET improves the network exploiting a single
alignment module by 0.4\% and 0.2\% respectively, as shown as Table \ref{ablationstudy}. The SSANET outperforms both the two subnetworks, which proves the scale alignments can improve the performance of spatial alignments with collaborative learning. By virtue of the DSM module and OP module, our part features enable to represent the whole image with no need for the global feature.

\paragraph{robustness to spatial and scale variance}
To validate that spatial and scale misalignment influence the performance dramatically, we rotate the images of CUB as the input. The performance of a raw ResNet50 drops from 84.5\% to 57.3\% while our SSANET drops from 88.5\% to 65.9\%, as shown in Table \ref{rotation}. And we did not rotate the data for augmentation during the training process. It manifolds that a raw CNN has poor invariance to rotation and SSANET can improve the situation significantly. Similarly, We reduce the image to a size of 300 $\times$ 300 and then pad it to the size of 448 $\times$ 448 as the input. The performance of a raw ResNet50 drops from 84.5\% to 76\% while our SSANET drops from 88.5\% to 83.2\%, as shown in Table \ref{shrink}. SSANET can enhance the robustness to scale invariance to a certain extent.
\begin{figure}
\begin{minipage}[w]{0.2\textwidth} 
\makeatletter\def\@captype{table}
\small

  \caption{Effect of OPM module vs. Global Average Pooling (GAP) vs. Global Max Pooling (GMP) based on ResNet-50 + MAC on CUB-200-2011.}
  \label{ablationonopm}
  \centering
  \begin{tabular}{ll}
  \hline
    \toprule
    Method        & Top-1 \\
    \midrule
  GMP        &86.0  \\    
  GAP        &87.3  \\
  OP   &88.1  \\
    \bottomrule
  \end{tabular}
\end{minipage} 
\quad
\begin{minipage}[w]{0.25\textwidth} 
\makeatletter\def\@captype{table}
\small
  \caption{Effect of DSM module vs. pyramid methods based on ResNet-50 + MAC on CUB-200-2011.}
  \label{ablationondsm}
  \centering
  \begin{tabular}{ll}
  \hline
    \toprule
    Method        & Top-1 \\
    \midrule
    SPPNet \cite{he2015spatial}   & 87.4 \\
    FPN \cite{lin2017feature}     & 87.6 \\
    DSM(fixed) &  88.0  \\
    DSM(1)     &  88.1  \\
    DSM        &  88.3  \\
    \bottomrule
  \end{tabular} 
\end{minipage} 
\end{figure}
\subsection{Comparison with state-of-the-art approaches}
\paragraph{Quantitative Results}
We evaluate the performance of the proposed SSANET against 9 state-of-the-art algorithms, including RACNN~\cite{fu2017look}, MACNN~\cite{zheng2017learning}, MAMC~\cite{sun2018multi}, MaxEnt~\cite{NIPS2018_7344}, DFL-CNN~\cite{wang2018learning}, NTSNet~\cite{yang2018learning}, HSnet~\cite{lam2017fine}, TASN\cite{zheng2019looking} and iSQRT-COV\cite{li2018towards}.
The results of different algorithms on the CUB-200-2011 dataset are presented in Table \ref{comparison1}. Overall, our proposed approach outperforms all previous methods.
Previous part-based methods can be divided into two methodologies in general. Part-aware methods like  \cite{zheng2017learning,sun2018multi} focus on channel attention module to distinguish different parts via diverse channels and supervise for each channel. Part-based methods like \cite{fu2017look,wang2018learning,yang2018learning} focus on capturing regions of interest (ROI) as marginal part information and supervise for each part regions.

All the above methods fuse global and complementary part features. It is worth mentioning that our method merely harnesses aligned part features to recognition image-level category due to more accurate part localization. Yet our method outperforms all of the 9 approaches  by 3.2\%, 2.0\%, 2.0\%, 2.0\%, 1.1\%, 1.0\%, 1.0\%, 0.6\%, 0.4\% respectively. In addition, our method outperforms the state-of-the-art part-annotated methods like HSnet \cite{lam2017fine} by 1.0\%, which proves that our proposed method enables to capture discriminative part localization without extra semantic navigation.

\begin{table}
\small
  \caption{Comparison results on CUB-200-2011. "Aux" stands
for using extra annotation in training. "Fusion" denotes the way of feature fusion. "G" represents global features and "P" represents part features.}
  \label{comparison1}
  \centering
\begin{tabular}{lllll}
    \toprule
    Method     & Backbone  &Aux  & Fusion  & Top-1 \\ 
    \midrule
    RACNN\cite{fu2017look}     &  VGG-19      &\ding{56}   & G+P & 85.3     \\
    MACNN\cite{zheng2017learning}    & VGG-19       &\ding{56}  & G+P  & 86.5     \\
    MAMC\cite{sun2018multi}   & ResNet-101   &\ding{56}  & G+P &  86.5      \\
    MaxEnt\cite{NIPS2018_7344}   & DenseNet-161 &\ding{56} & G  & 86.5  \\
    DFL-CNN\cite{wang2018learning}  & ResNet-50  &\ding{56}   & G+P   & 87.4  \\
    NTSNet\cite{yang2018learning}  & ResNet-50    &\ding{56} & G+P   & 87.5  \\
    HSnet\cite{lam2017fine}  &GoogLeNet     & \checkmark & G+P  & 87.5 \\
    iSQRT-COV \cite{li2018towards} &ResNet-50    & \ding{56} & G  & 88.1 \\
    TASN\cite{zheng2019looking}    &ResNet-50    & \ding{56} & G+P  & 87.9 \\
    \midrule
    SSANET  & ResNet-50  &\ding{56} & P & \boldmath${88.5}$ \\
    SSANET  & ResNet-101  &\ding{56} & P & \boldmath${88.7}$ \\
    \bottomrule
  \end{tabular}
\end{table}


Our method obtains new state-of-the-art performances on the Stanford Cars and FGVC Aircraft datasets, as shown in Table \ref{car&air}. Our model achieves the ${94.6\%}$ and ${92.3\%}$ top-1 accuracy on the Stanford Cars dataset and FGVC Aircraft dataset respectively.

\begin{figure}[htbp]
\centering
\includegraphics[width=0.38\textwidth]{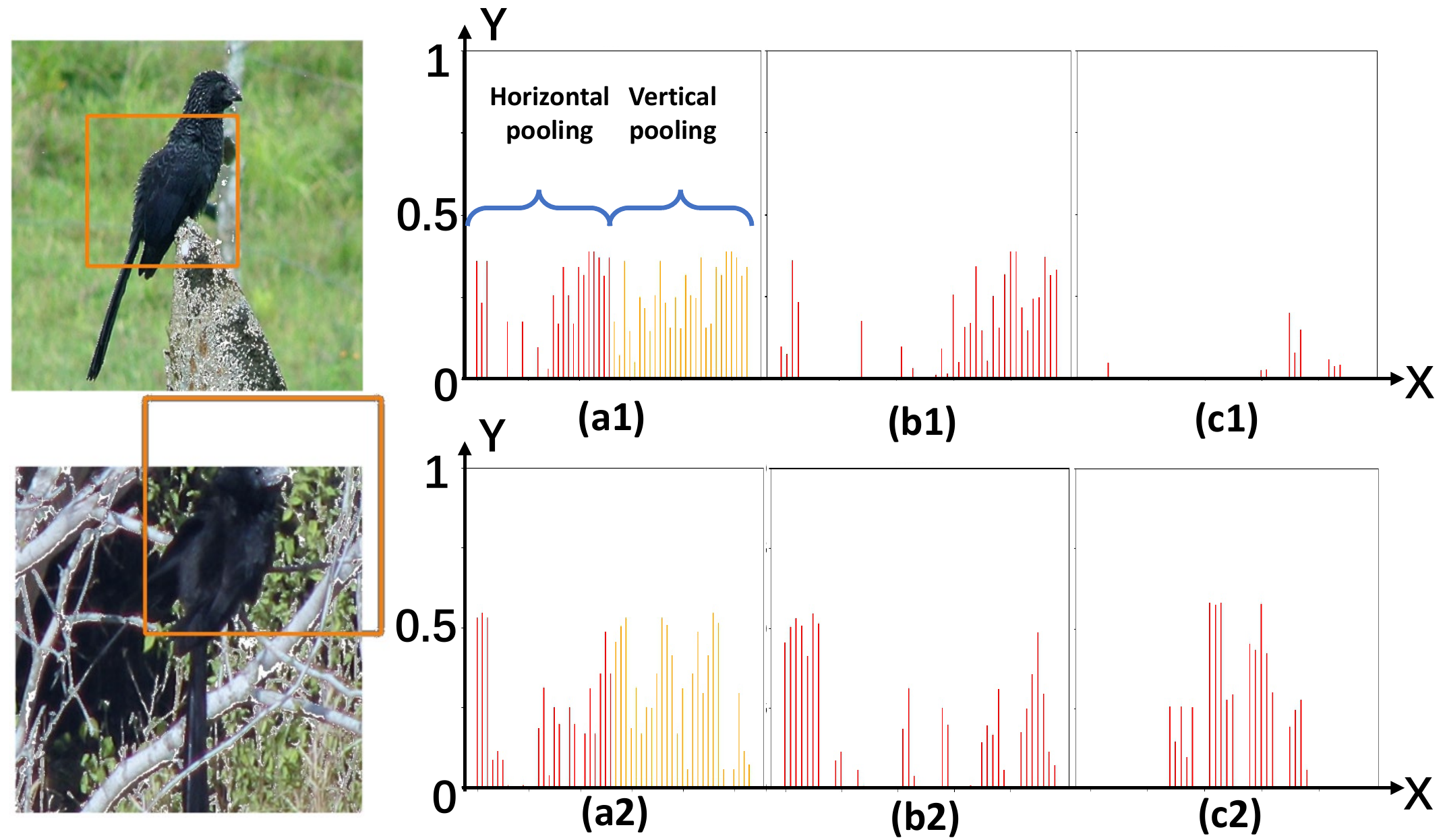}
\caption{Qualitative results of the OP module for the CUB dataset. The left images show the ROI regions on the original images (e.g. the trunk of birds). ($a1,a2$) are the corresponding histograms in which X-axis means the flatten vector of two oriented poolings (horizontal pooling and vertical pooling) and Y-axis means the quantitative value. The middle histograms ($b1$,$b2$) show the quantized value of flatten $7 \times 7$ feature maps trained with the OP module, while the right histogram ($c1$,$c2$) are the results without the OP module.}
\label{alignment_result}
\end{figure}

\begin{figure}
\centering
\includegraphics[width=0.45\textwidth]{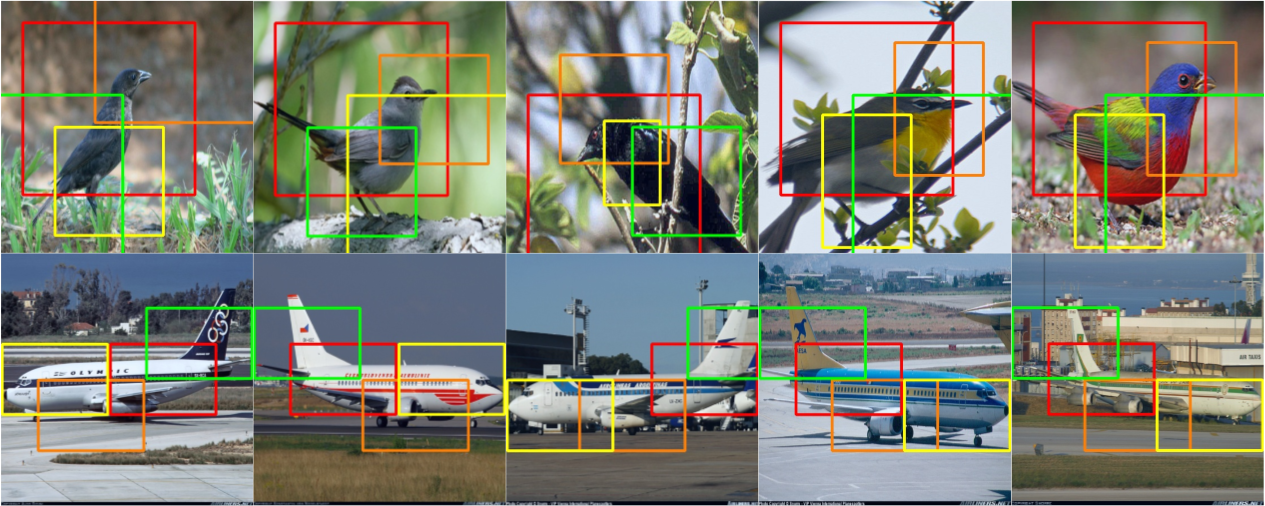}
\caption{Qualitative results of SSANET for birds and airplanes.}
\label{bird_airplane}
\end{figure}
\paragraph{Qualitative Results}
We qualitatively shows the our learned part localizations compared with NTSNet \cite{yang2018learning} and the ground truth. With the morphological alignment constraints, our learned proposal regions usually correspond to the semantically informative regions.  For instance, our $1st$ proposal (green boxes in the $1st$ row, $2nd,3th$ column) refer to the eyes of birds and $2nd$ proposal (orange boxes in the $2nd$ row, $2nd,3th$ column) refer to the wings. As a contrast, the proposals of NTSNet ($1st$ column) are more chaotic, failing to align semantic correlations.
Profiting from our DSM module, the center of our proposals are more adjacent to GT compared with that of NTSNet though we train SSANET without extra annotations like part key points.
Considering scale estimation, the ROIs of SSANET stick to the principal part tightly, avoiding vital feature missing or redundant context.
In addition, with the proposed OP module,  we can not only align the orientation and rotation variance, but also strength the insensitivity of background noise (Figure \ref{alignment_result}). For the same ROI of two images in the same class, we can observe that the distributions of $7 \times 7$ feature map ($b1,b2$) are much more similar of two intra-class objects while the distributions of original $7 \times 7$ feature map without the OP module ($c1,c2$) are with huge difference.  Also, we can see that the distribution of both horizontal pooling and vertical pooling among the same-class images are similar, which manifolds the effectiveness of our OP module.
Moreover, we draw the generated regions predicted by SSANET, as shown in Figure \ref{bird_airplane}. By Morphgical Alignment Constraints, the key regions, e.g., the head and wing for birds, the tail for aircrafts, can be highlighted and aligned.

\begin{table}
\small
  \caption{Comparison in terms of classification accuracy on the FGVC Aircraft dataset.}
\label{car&air}
  \centering
  \begin{tabular}{lll}
    \toprule
    \cmidrule(r){1-3}
    Method     & Backbone     & Top-1 \\
    \midrule
    MACNN\cite{zheng2017learning} & VGG-19  & 89.9     \\
    MaxEnt\cite{NIPS2018_7344}   & DenseNet-161    &  89.8 \\
    DFL-CNN\cite{wang2018learning}   & VGG-16     & 92  \\
    iSQRT-COV\cite{li2018towards} &ResNet-50    & 90.0 \\
    NTSNet\cite{yang2018learning}    & ResNet-50       & 91.4  \\
    \midrule
    SSANET   & ResNet-50      & \boldmath${92.3}$ \\
    \bottomrule
  \end{tabular}
\end{table}

\begin{table}
\small
  \caption{Comparison in terms of classification accuracy on the Stanford Cars dataset.}
\label{car&air}
  \centering
  \begin{tabular}{lll}
    \toprule
    \cmidrule(r){1-3}
    Method     & Backbone     & Top-1 \\
    \midrule
    RACNN\cite{fu2017look} &  VGG-19  & 92.5     \\
    MACNN\cite{zheng2017learning} & VGG-19  &  92.8   \\
    MAMC\cite{sun2018multi}    & ResNet-101  & 93.0       \\
    MaxEnt\cite{NIPS2018_7344}    & ResNet-50    & 93.9  \\
    DFL-CNN\cite{wang2018learning}   & VGG-16     &  93.8 \\
    NTSNet\cite{yang2018learning}    & ResNet-50       & 93.9  \\
    HSnet\cite{lam2017fine}   & GoogLenet &93.9 \\
    iSQRT-COV\cite{li2018towards} &ResNet-50    & 92.8 \\
    TASN \cite{zheng2017learning}   &ResNet-50     & 93.8 \\
    iSQRT-COV\cite{li2018towards} &ResNet-50    & 92.8 \\
    \midrule
    SSANET    & ResNet-50      & \boldmath${94.6}$ \\
    \bottomrule
  \end{tabular}
\end{table}

\section{Conclusion}
In this paper, we propose a spatial-scale aligned network to learn spatial-scale invariance jointly for part regions. The proposed network does not need bounding box/annotations for training but can be qualified to obtain accurate part localizations as well as spatial-scale estimation. Extensive experiments demonstrate the state-of-the-art performance on both multiple-part localization and recognition on birds, automobiles, and airplanes.



{\small
\bibliographystyle{ieee_fullname}
\bibliography{references}
}
\end{document}